\documentclass[sigconf]{acmart}

\AtBeginDocument{%
  \providecommand\BibTeX{{%
    \normalfont B\kern-0.5em{\scshape i\kern-0.25em b}\kern-0.8em\TeX}}}

\setcopyright{acmcopyright}
\copyrightyear{2023}
\acmYear{2023}
\setcopyright{acmlicensed}\acmConference[WWW '23]{Proceedings of the ACM Web Conference 2023}{May 1--5, 2023}{Austin, TX, USA}
\acmBooktitle{Proceedings of the ACM Web Conference 2023 (WWW '23), May 1--5, 2023, Austin, TX, USA}
\acmPrice{15.00}
\acmDOI{10.1145/3543507.3583241}
\acmISBN{978-1-4503-9416-1/23/04}

\usepackage{algorithm}
\usepackage{algorithmic}

\usepackage{amsmath}
\usepackage{subcaption}

\usepackage{amsmath}
\usepackage{graphicx}
\usepackage{graphics}
\usepackage{multirow}

\newcommand{\mname}{DPAO}

\begin{document}

\title{Dual Policy Learning for Aggregation Optimization in \\Graph Neural Network-based Recommender Systems}

\author{Heesoo Jung}
\affiliation{
  \institution{Sungkyunkwan University}
  \city{Suwon}
  \country{Republic of Korea}
}
\email{steve305@skku.edu}

\author{Sangpil Kim}
\affiliation{
  \institution{Korea University}
  \city{Seoul}
  \country{Republic of Korea}
}
\email{spk7@korea.ac.kr}

\author{Hogun Park}\authornote{Corresponding Author}
\affiliation{
  \institution{Sungkyunkwan University}
  \city{Suwon}
  \country{Republic of Korea}
}
\email{hogunpark@skku.edu}

\begin{abstract}
  Graph Neural Networks (GNNs) provide powerful representations for recommendation tasks. 
  GNN-based recommendation systems capture the complex high-order connectivity between users and items by aggregating information from distant neighbors and can improve the performance of recommender systems.
  Recently, Knowledge Graphs (KGs) have also been incorporated into the user-item interaction graph to provide more abundant contextual information; they are exploited to address cold-start problems and enable more explainable aggregation in GNN-based recommender systems (GNN-Rs). 
However, due to the heterogeneous nature of users and items, developing an effective aggregation strategy that works across multiple GNN-Rs, such as LightGCN and KGAT, remains a challenge.
  In this paper, we propose a novel reinforcement learning-based message passing framework for recommender systems, which we call \mname{} (\textit{\textbf{D}ual \textbf{P}olicy framework for \textbf{A}ggregation \textbf{O}ptimization}). 
  This framework adaptively determines high-order connectivity to aggregate users and items using dual policy learning. 
  Dual policy learning leverages two Deep-Q-Network models to exploit the user- and item-aware feedback from a GNN-R and boost the performance of the target GNN-R. 
  Our proposed framework was evaluated with both non-KG-based and KG-based GNN-R models on six real-world datasets, and their results show that our proposed framework significantly enhances the recent base model, improving $nDCG$ and $Recall$ by up to 63.7\% and 42.9\%, respectively. Our implementation code is available at \url{ https://github.com/steve30572/DPAO/}.
\end{abstract}



\maketitle


\section{Introduction}
Recommender systems have been used in various fields, such as E-commerce and advertisement. The goal of recommender systems is to build a model that predicts a small set of items that the user may be interested in based on the user's past behavior. Collaborative Filtering (CF)~\cite{CF1, CF2} provides a method for personalized recommendations by assuming that 
similar users would exhibit analogous preferences for items.
Recently, Graph Neural Networks (GNNs)~\cite{gcn1,gcn2} have been adapted to enhance recommender systems~\cite{NGCF,LightGCN} by refining feature transformation, neighborhood aggregation, and nonlinear activation.  
Furthermore, Knowledge Graphs (KGs) are often incorporated in GNN-based recommender systems~(GNN-Rs)~\cite{kgat,ckan,kgin} as they provide important information about items and users, helping to improve performance in sparsely observed settings.
GNN-Rs have demonstrated promising results in various real-world scenarios, and GNNs have become popular architectural components in recommendation systems.


The aggregation strategy in GNN-based recommendation systems (GNN-Rs) is crucial in capturing structural information from the high-order neighborhood. For example, LightGCN~\cite{LightGCN} alters the aggregation process by removing activation functions and feature transformation matrices. KGAT~\cite{kgat} focuses on various edge types in the Collaborative Knowledge Graph (CKG) and aggregates messages based on attention values.
However, many current GNN-Rs-based approaches have limitations in terms of fixing the number of GNN layers and using a fixed aggregation strategy, which restricts the system's ability to learn diverse structural roles and determine more accurate ranges of the subgraphs to aggregate. This is particularly important because these ranges can vary for different users and items. For instance, while one-hop neighbors only provide direct collaborative signals, two or three-hop neighbors can identify everyday consumers or vacation shoppers when knowledge entities are associated.

\begin{figure}[t]
  \centering
  \includegraphics[width=\linewidth]{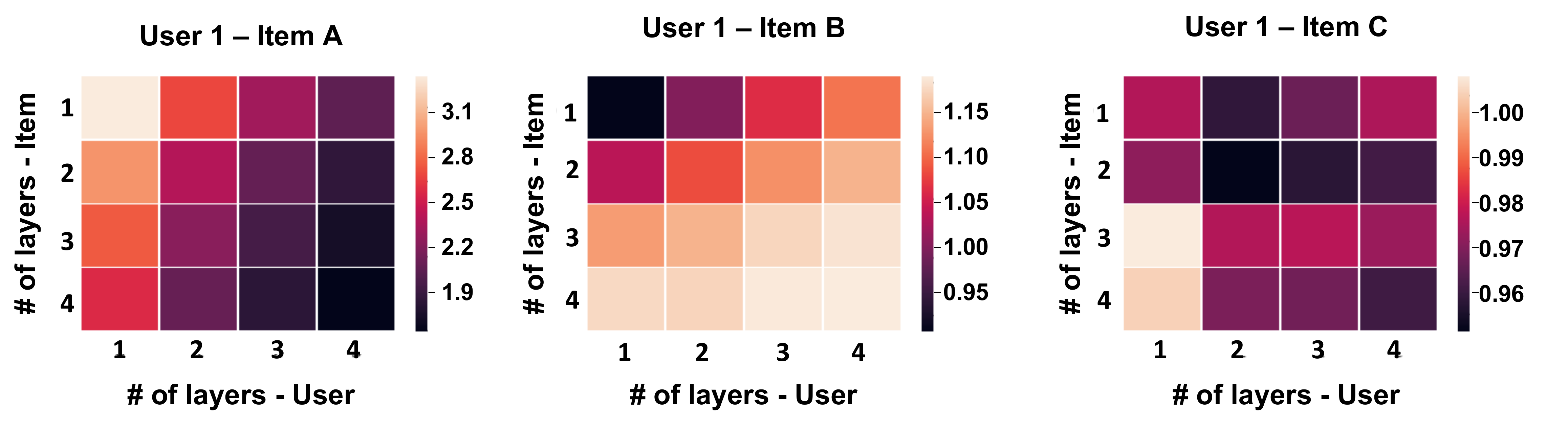}
  \caption{The effect of different iterations of aggregations in LightGCN~\cite{LightGCN} on MovieLens1M dataset. The X-axis denotes the number of aggregation layers for its user-side GNNs. The Y-axis denotes the number of aggregation layers for item-side GNNs.
  The color from dark to bright represents the matching score of the user and the item, which is computed from the inner product of the user and item representations.}
  \label{fig:intro}
  \vspace{-4 mm}
\end{figure}

In this context, we hypothesize that each user and item may benefit from adopting a different aggregation strategy of high-order connectivity to encode better representations, while most current models fix the number of GNN layers as a hyperparameter.
To demonstrate the potential significance of this hypothesis, we experiment with LightGCN~\cite{LightGCN} by varying the number of layers in the user-side and item-side GNNs.
The matching scores of a sampled user (User 1) with three interacted items (Item A, Item B, and Item C) in the MovieLens1M dataset are illustrated in Figure~\ref{fig:intro}.
The matching score shows how likely the correct item is to be predicted, and a brighter color indicates a higher score.
The first sub-figure in Figure~\ref{fig:intro} indicates that one item-side GNN layer and one user-side GNN layer capture the highest matching score (i.e., the highest value at location (1,1) in the heat map).
However, the second and third subfigures show that the (4,4) and (3,1) locations of the heat map have the highest scores, respectively.
These observations inspired us to develop a novel method that adaptively selects different hops of neighbors, which can potentially increase the matching performance and learn different selection algorithms for users and items because of the heterogeneous nature of users and items.

In this paper, we propose a novel method that significantly improves the performance of GNN-Rs by introducing an adaptive aggregation strategy for both user- and item-side GNNs.
We present a novel reinforcement-learning-based message passing framework for recommender systems, \mname{}~(\textit{\textbf{D}ual \textbf{P}olicy framework for \textbf{A}ggregation \textbf{O}ptimization}). 
\mname{} newly formulates the aggregation optimization problem for each user and item as a Markov Decision Process (MDP) and develops two Deep-Q-Network (DQN) \cite{DQN} models to learn Dual Policies. Each DQN model searches for effective aggregation strategies for users or items.
The reward for our MDP is obtained using item-wise sampling and user-wise sampling techniques.

The contributions of our study are summarized as follows:
\begin{itemize}
    \item We propose a reinforcement learning-based adaptive aggregation strategy for a recommender system.
    \item We highlight the importance of considering the heterogeneous nature of items and users in defining adaptive aggregation strategy by suggesting two RL models for users and items.
    \item \mname{} can be applied to many existing GNN-R models that obtain the final user/item embedding from layer-wise representation.
    \item We conducted experiments on both non-KG-based and KG-based datasets, demonstrating the effectiveness of \mname{}.
\end{itemize}

\begin{figure*}[t]
  \centering
  \includegraphics[width=\linewidth]{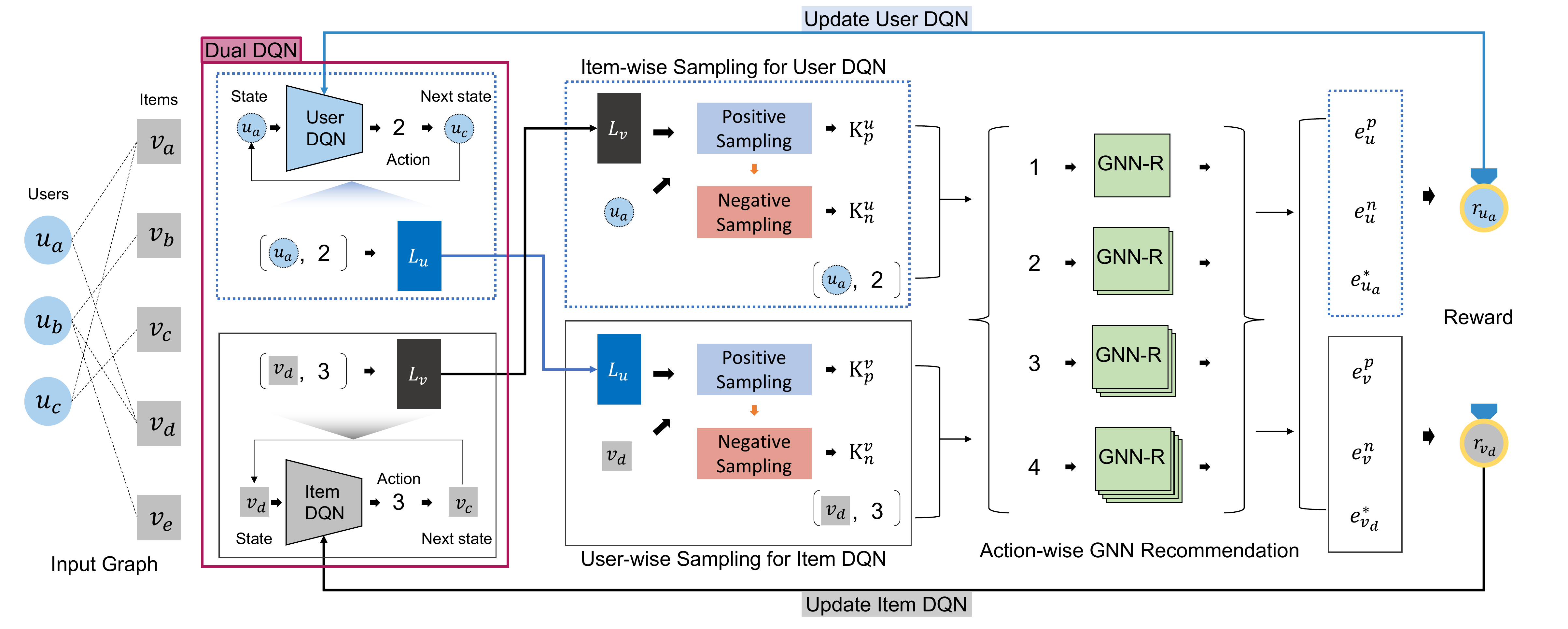}
  \vspace{-6 mm}
  \caption{Illustration of the overall flow of~\mname{}. 
  The left side of the figure is the user-item bipartite input graph.
  Then it shows the MDP of user DQN (colored blue) and item DQN (colored grey).
  We perform item-wise sampling for user DQN and user-wise sampling for item DQN.
  Followed by the action-wise GNN-R, we achieve the final embeddings.
  Reward values acquired from the embeddings are also used to update our dual DQN.}
  \label{fig:Traversing}
  \vspace{-2 mm}
\end{figure*}


\section{Preliminaries}

\subsection{Graph Neural Networks in Recommender System (GNN-R)}
\label{sec:GNN-R}
We define a user-item bipartite graph $(\mathcal{G})$ that includes a set of users $\mathcal{U} = \{u\}$ and items $ \mathcal{V} = \{v\}$.  In addition, users' implicit feedback $\mathcal{O}^{+} = \{(u, v^{+})|u \in \mathcal{U}, v^{+} \in \mathcal{V}\}$ indicates a interaction between user $u$ and positive item $v^+$.
We define $I$ as the interaction set.
The interaction set for user $u$ is represented by $I(u) = \{v | v \in \mathcal{V},  (u, v) \in \mathcal{O}^{+}\}$.
For a recommender system using KG, a Knowledge Graph is composed of Entity-Description~($\mathcal{D}$) - Relation ($\mathcal{R}$) - Entity-Description~($\mathcal{D}$) triplets $\{(h,r,t)|h,t \in \mathcal{D}, r \in \mathcal{R}\}$ indicating that the head entity description $h$ is connected to the tail entity description $t$ with relation $r$. 
An entity description is also known as an entity.
Items in a bipartite graph are usually an entity of the KG.
Therefore, a Collaborative Knowledge Graph (CKG) is constructed seamlessly by combining KG and user-item bipartite graph using the KG's item entity~\cite{kgat, kgin}.

The GNN-R model is expressed in three steps: aggregation, pooling, and score function, which are described in detail as follows:

\textbf{Aggregation.} First, each user $u$ and item $v$ are represented as the initial embedding vectors $\mathbf{e_u^0}$ and $\mathbf{e_v^0}$, respectively. 
An aggregation function in GNN-Rs aims to encode the local neighborhood information of $u$ and $v$ to update their embedding vectors. 
For example, the aggregation function of LightGCN~\cite{LightGCN} can be expressed as
\begin{equation}
\begin{aligned}
    \mathbf{e}_u^{(n+1)} = \sum_{j\in N_u} \frac{1}{\sqrt{|\mathcal{N}_u|}\sqrt{|\mathcal{N}_j|}} \mathbf{e}_j^{(n)} , \\
    \mathbf{e}_v^{(n+1)} = \sum_{k\in N_v} \frac{1}{\sqrt{|\mathcal{N}_v|} \sqrt{|\mathcal{N}_k|}} \mathbf{e}_k^{(n)},
    \label{equ:aggregation}
\end{aligned}
\vspace{-2mm}
\end{equation}

\noindent where $\mathbf{e}_u^{(n)}$ and $\mathbf{e}_v^{(n)}$ are embedding vectors of $u$ and $v$ at the $n^{\text{th}}$ layer, respectively. 
$\mathcal{N}_u$ and $\mathcal{N}_v$ denotes the neighbor sets of $u$ and $v$, respectively.
Each GNN-layer aggregates the local neighborhood information and then passes this aggregated information to the next layer. Stacking multiple GNN layers allows us to obtain high-order neighborhood information. 
A specific number of GNN layers affects the recommendation performance~\cite{NGCF, LightGCN}.

\textbf{Pooling.} The aggregation functions of GNN-R models produce user/item embedding vectors in each layer. 
For example, assuming that $(N)$ GNN layers exist, $(N+1)$ user embeddings ($\mathbf{e}_u^0, \mathbf{e}_u^1, ... , \mathbf{e}_u^N$) and $(N+1)$ item embeddings ($\mathbf{e}_v^0, \mathbf{e}_v^1, ... , \mathbf{e}_v^N$) are returned. 
To prevent over-smoothing~\cite{oversmoothing}, pooling-based approaches are often leveraged to obtain the final user and item representations. Two representative methods are summation-based pooling~\cite{LightGCN, sum1} and concatenation-based pooling~\cite{NGCF, DGCF, kgin}. 
For example, summation-based pooling can be represented as
\begin{equation}
    \mathbf{e_u^{*}} = \sum_{n=0}^{N}\lambda_n \mathbf{e_u^{(n)}}, \qquad
    \mathbf{e_v^{*}} = \sum_{n=0}^{N}\lambda_n \mathbf{e_v^{(n)}},
    \label{equ:sumpool}
\end{equation}
\noindent where $\mathbf{e_u^*}$ and $\mathbf{e_v^*}$ represent the final embedding vectors of $u$ and $v$, respectively. 
$\lambda_n$ denotes the weight of the $n^{\text{th}}$ layer. Concatenation-based pooling concatenates all $(N+1)$ embedding vectors to represent the final vectors.

\textbf{Score Function.} By leveraging the final user/item representations above, a (matching) score function returns a scalar score to predict the future interaction between user $u$ and item $v$. 
For example, a score function between $u$ and $v$ is given as
\begin{equation}
    Score(\mathbf{e_u^*}, \mathbf{e_v^*}) = \mathbf{e_u^*}^T \mathbf{e_v^*}.
    \label{equ:score}
\end{equation}

\textbf{Loss Function.} Most loss functions in GNN-Rs use BPR loss~\cite{bprloss}, which is a pairwise loss of positive and negative interaction pairs. The loss function can be expressed as
\begin{equation}
\mathcal{L}_{GNN} = \sum_{i \in I(u), j \notin I(u)} -\ln{\sigma(Score(\mathbf{e_u^{*}},\mathbf{e_i^{*}}) - Score(\mathbf{e_u^{*}}, \mathbf{e_j^{*}}))}, 
\label{equ:lossfunction}
\vspace{-2mm}
\end{equation}
\noindent where $\sigma$ is an activation function. ($u$, $i$) is an observed interaction pair in an interaction set $I$ and ($u$, $j$) is a negative interaction pair that is not included in $I$.

\subsection{Reinforcement Learning (RL)}
\textbf{Markov Decision Process.} The Markov Decision Process (MDP) is a framework used to model decision-making problems in which the outcomes are partly random and controllable. 
The MDP is expressed by a quadruple $(\mathcal{S, A},  \mathcal{P}_{t}, \mathcal{W})$, where $\mathcal{S}$ is a state set, $\mathcal{A}$ is an action set, and $\mathcal{P}_{t}$ is the probability of a state transition at timestamp $t$. 
$\mathcal{W}$ is the reward set provided by the environment. 
The goal of MDP is to maximize the total reward and find a policy function $\pi$. The total reward can be expressed as $\mathbb{E}_\pi[\sum_{t=0}^\infty \gamma^t \cdot r_t]$, where $r_t$ is the reward given at timestamp $t$, and $\gamma$ is the discount factor of reward. The policy function $\pi$ returns the optimum action that maximizes the cumulative reward.

Deep Reinforcement Learning (DRL) algorithms solve the MDP problem with neural networks~\cite{pg} and identify unseen data only by observing the observed data~\cite{DRL1}.
Deep Q-Network (DQN)~\cite{DQN}, which is one model of the DRL, approximates Q values with neural network layers. The Q value is the estimated reward from the given state $s_t$ and action $a_t$ at timestamp $t$. $Q(s_t,a_t)$ can be expressed as
\begin{equation}
    Q(s_t,a_t) = \mathbb{E}_\pi[r_t + \gamma\max_{a}Q(s_{t+1}, a_{t+1})],
    \label{equ:Qoptim}
\end{equation}
where $s_{t+1}$ and $a_{t+1}$ are the next state and action, respectively.
The input of the DQN is a state, and the output of the model is a distribution of the Q value with respect to the possible actions.

However, it is difficult to optimize a DQN network owing to the high sequence correlation between input state data.
Techniques such as a replay buffer and separated target network~\cite{DQN1} help to stabilize the training phase by overcoming the above limitations.


\section{Methodology}

This section presents our proposed model, \textit{\textbf{D}ual \textbf{P}olicy framework for \textbf{A}ggregation \textbf{O}ptimization}~(\mname{}).
Our Dual Policy follows a Markov Decision Process~(MDP), as shown in Figure~\ref{fig:Traversing}.
It takes initially sampled user/item nodes ($u_a, v_d$ in Figure~\ref{fig:Traversing}) as states and maps them into actions (i.e., the number of hops) of both the user and item states (two and three, respectively, in Figure~\ref{fig:Traversing}).
In other words, the actions of the user or item states determine how many of the corresponding GNN layers are leveraged to obtain the final embeddings and compute user-item matching scores. 
Every (state, action) tuple is accumulated as it proceeds.
For example, ($u_a, 2$) is saved in the user's tuple list $L_u$, and ($v_d, 3$) is assigned to $L_v$.
From the given action, it stacks action-wise GNN-R layers and achieves the user's reward via item-wise sampling and the item's reward with user-wise sampling.
Our proposed framework could be integrated into many GNN-R models, such as LightGCN \cite{LightGCN}, NGCF \cite{NGCF}, KGAT \cite{kgat}, and KGIN \cite{kgin}. 

In the following section, we present the details of \mname{}. 
We first describe the key components of MDP: state, action, and next state.
Then, we introduce the last component of MDP, reward.
Next, we discuss the time complexity of \mname{} in detail.
Finally, the integration process to GNN-R models is shown.

\subsection{Defining MDP in Recommender System}

\begin{figure}[t]
  \centering
  \includegraphics[width=\linewidth]{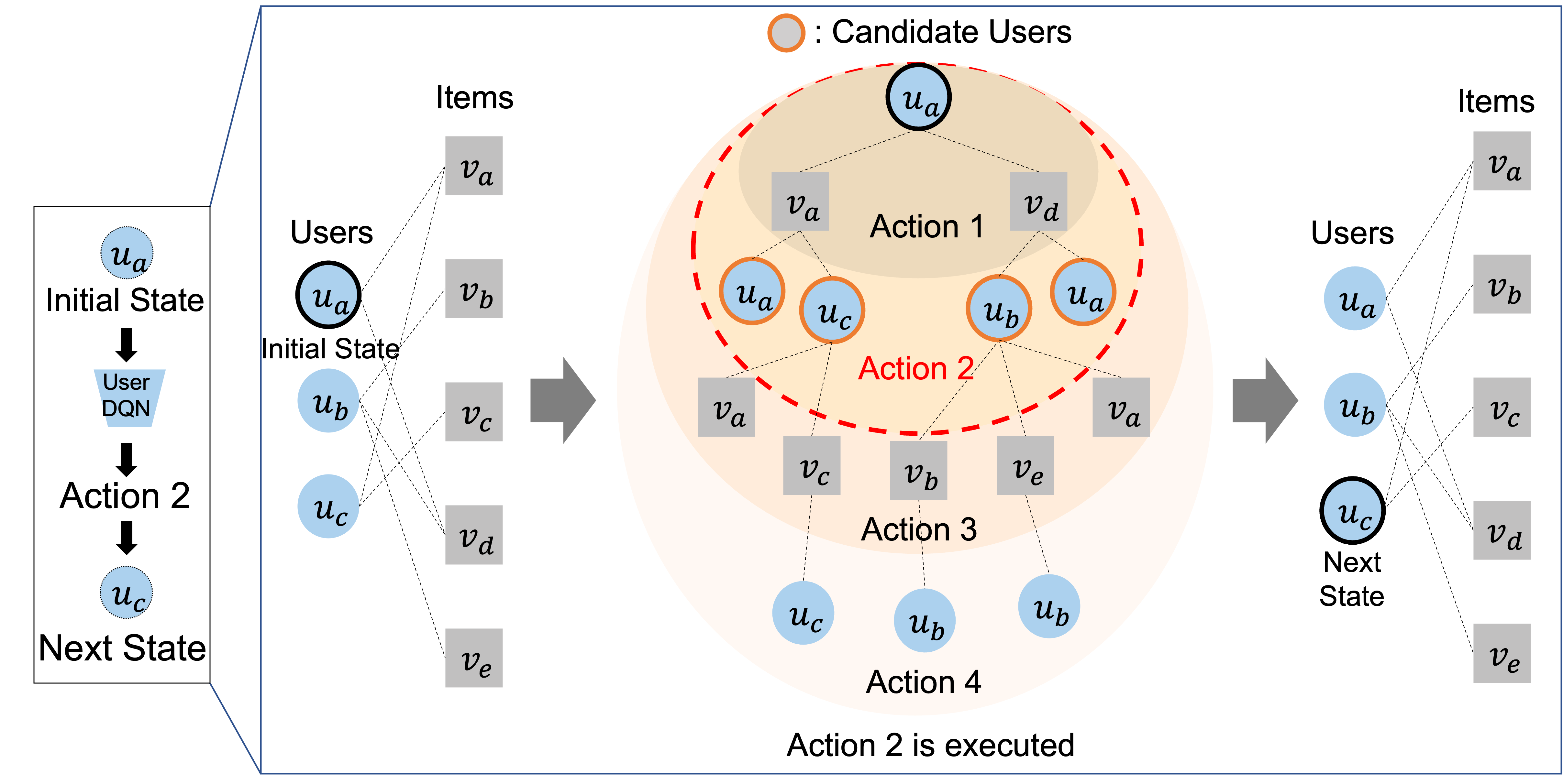}
  
  \caption{The procedure of acquiring next state on \mname{}. 
If the action of its timestamp is two, then the state $u_a$ is transited to $u_c$, chosen from the two-hop neighbors. }
  \label{fig:next}
  \vspace{-2mm}
\end{figure}

The goal of our MDP is to select an optimal aggregation strategy for users and items in a recommender system. 
Two different MDPs are defined to differentiate the policies for users and items as follows:

\begin{algorithm}[!t]
\caption{\textit{\textit{NextState}}($s_t, a_t, \mathcal{G}$) // Acquisition of the next state }\label{alg:next}
\begin{algorithmic}[1]
\STATE{\textbf{Input}: state $s_t$, action $a_t$ and input graph $\mathcal{G}$}
\STATE{\textbf{Output}: $s_{t+1}$}
\STATE{Get the ($a_t$)-hop subgraph $\mathcal{G}_{s_t}^{a_t}$ from  $\mathcal{G}$ around $s_t$}
\STATE{$\mathcal{M}= \{m| m\in G_{s_{t}}^{a_t}, type(s_t)==type(m)\}$}
\STATE \COMMENT{$type(s_t)$ returns the node type of $s_t$, which can be a user or an item.}
\STATE{\textbf{return} a node $m \in \mathcal{M}$ with uniform random sampling}

\end{algorithmic}
\end{algorithm}

\paragraph{\textbf{User's MDP}}  A state set $\mathcal{S}$ represents user nodes in an input graph. The initial state of user $s_0^u$ is randomly selected from the users in the input graph. 
An action selects the number of hops from the action set $\mathcal{A}$ that maximizes its reward. 
The state and action information is saved to the tuple list $L_u$.
With state $s_t^u$ and action $a_t^u$ at timestamp $t$, 
the next state can be obtained.
Algorithm~\ref{alg:next} provides details of the acquisition of the next state.
With $s_t^u$ and $a_t^u$, the $a_t^u$-hop subgraph around $s_t^u$,  $\mathcal{G}_{s_t^u}^{a_t^u}$, can be generated (Line(L)~3 in Alg.~\ref{alg:next}).
As the nodes in $\mathcal{G}_{s_t^u}^{a_t^u}$ are nodes where the target node ($s_t^u$) receives information, we choose the next state from the nodes in this subgraph.
Thus, the sampling space for the next state $s_{t+1}^u$ is determined. 
Subsequently, a candidate set $\mathcal{M}$ is defined from the subgraph (L~4 in Alg.~\ref{alg:next}), and the next user state is chosen with a uniform transition probability $\frac{1}{|\mathcal{M}|}$. 
Figure~\ref{fig:next} shows an example of state transition. Assume that the current state is at $u_a$ and the chosen action is 2.
The candidate user set $\mathcal{M}$ is then determined to be $\{u_a, u_c, u_b\}$.
$u_c$ is randomly selected and becomes the next user state.

\paragraph{\textbf{Item's MDP}} Similarly, with the user's MDP, the item's MDP can be defined.
The only difference from the user's MDP is that the state set of the item's MDP comprised the item nodes. 
Moreover, the initial state $s_0^v$ is chosen at the interaction set of the initial user state $I(s^u_0)$.
The state and action information of the state of the item are stored in tuple list $L_v$.

\begin{algorithm}[t]
\caption{\textit{Reward}($s_t, a_t, L, I, \theta_{gnn}$)}\label{alg:reward}
\begin{algorithmic}[1]
\STATE{\textbf{Input:} state $s_t$, action $a_t$, list of state/action tuples $L (L_u$ or $L_v)$, interaction set $I$, and GNN-R parameters $\theta_{gnn}$}
\STATE{\textbf{Output:} reward $r_t$ of $s_t$}
\STATE{$\mathcal{K}_p = \{(s, a) \in L, s = s_i \}$ for all $s_i \in I(s_t)$}
\STATE{$\mathcal{K}_n = \{(s, a) \in L, s \neq s_i \}$ for all $s_i \in I(s_t)$}
\STATE{Sample $|\mathcal{K}_p|$ number of tuples from $\mathcal{K}_n$ and Store them into $\mathcal{K}'_n$}
\STATE{Get $\mathbf{e}_{s_t}^{*}$ from $\text{GNN-R}_{\theta_{gnn}}(s_t, a_t)$}
\STATE{Get $\mathbf{e}_{t}^{p}$ from Mean-Pooling($\text{GNN-R}_{\theta_{gnn}}(\mathcal{K}_p)$})
\STATE{Get $\mathbf{e}_{t}^{n}$ from Mean-Pooling($\text{GNN-R}_{\theta_{gnn}}(\mathcal{K}'_n)$})
\STATE{Compute $r_t$ using $\mathbf{e}_{s_t}^{*}$, $\mathbf{e}_{t}^{p}$, and $\mathbf{e}_{t}^{n}$ via Eq. \ref{equ:reward}}
\STATE{\textbf{return} $r_t$}
\end{algorithmic}
\end{algorithm}
\paragraph{\textbf{Reward of Both MDPs}}
The reward leverages the accumulated tuple list $L_u$ or $L_v$.
Therefore, the reward is calculated after a certain epochs (warm-up) while waiting for the tuple lists to be filled.
The opposite tuple list is leveraged to account for the heterogeneous nature of users and items.
The reward of users, which is related to the performance of the recommendation, needs to take into account the heterogeneous nature of item nodes.
In other words, $L_u$ is used to calculate the reward of the items, and $L_v$ is used to calculate the reward of users.
Algorithm~\ref{alg:reward} provides a detailed description of this process.
A positive-pair tuple set $\mathcal{K}_p$ is defined by finding positive interaction nodes not only in the tuple list $L$ but also in the given interaction set (L~3 in Alg.~\ref{alg:reward}).
Similarly, a negative-pair tuple set $\mathcal{K}_n$ finds the states in our tuple list $L$ that is not in our interaction set $I(s_t)$ (L~4 in Alg.~\ref{alg:reward}).
We randomly sample as many tuples on $\mathcal{K}_n$ as the length of $\mathcal{K}_p$ (L~5 in Alg.~\ref{alg:reward}).
The final representation vectors of the given state $e_{s_t}^{*}$ are obtained using action-wise GNN-R, $\text{GNN-R}_{\theta_{gnn}}(s_t, a_t)$. 
Correspondingly, the final representation vectors of positive pairs $\mathbf{e}_t^p$ and negative pairs $\mathbf{e}_t^n$ are acquired by applying the mean pooling of all the embedding vectors earned from $\mathcal{K}_p$ and $\mathcal{K}_n$ (L~7,~8 in Alg.~\ref{alg:reward}). 
Finally, the rewards for the user and item at timestamp $t$ are as follows:
\vspace{4mm}
\begin{equation}
\begin{aligned}
    r_t &= \frac{Score(\mathbf{e}_{s_t}^{*}, \mathbf{e}_{t}^{p}) - Score(\mathbf{e}_{s_t}^{*}, \mathbf{e}_{t}^{n})}{N(s_t, \mathbf{e}_t^{p}, \mathbf{e}_t^{n})},\\
    N(s_t, \mathbf{e}_t^p, \mathbf{e}_t^n) &= \sum_{c \in \mathcal{A}} (Score(\text{GNN-R}_{\theta_{gnn}}(s_t, c), e_t^p) \\
    & \ \ \ \ \ \ - \sum_{c \in \mathcal{A}}Score(\text{GNN-R}_{\theta_{gnn}}(s_t, c), e_t^n)), \\
    \label{equ:reward}
\end{aligned}
\vspace{-2mm}
\end{equation}
where $\mathbf{e}_{t}^{p}$ and $\mathbf{e}_{t}^{n}$ indicate the final representation of the positive and negative pairs, respectively.
The $Score$ function used in the numerator of the reward is given by Equation~(\ref{equ:score}).
The numerator of the reward follows the BPR loss (Equation (4)), which reflects the score difference in the dot products of the positive pair $(\mathbf{e}_{s_t}^{*}, \mathbf{e}_{t}^{p})$ and negative pair $(\mathbf{e}_{s_t}^{*}, \mathbf{e}_{t}^{n})$. 
Therefore, high numerator value results in similar representations with positive interactions. 
Moreover, it is important to measure the impact of the current positive pairs by considering the possible actions. 
The scale of the final reward should be adjusted by adding a normalization factor to the denominator.
The user's reward $r_t^u$ at timestamp $t$ is calculated through $Reward(s_t^u, a_t^u, L_v, I, \theta_{gnn})$.

An important part of acquiring the reward is that positive and negative pairs are sampled from our saved tuple lists, $L_u$, and $L_v$. The nodes stored in our tuple lists are indeed not far from their current state node because they are determined from the (action-hop) subgraph of the previous node. 
The negative samples that are close to the target node result in the preservation of its efficiency and the acquisition of stronger evidence for determining the accurate decision boundary.
Therefore, we assume that nodes in $\mathcal{K}_n$ can provide hard negative examples, which can enforce the optimization algorithm to exploit the decision boundary more precisely with a small number of examples.

\subsection{Optimizing Dual DQN}

To find an effective aggregation strategy in the action spaces in the MDPs above, DQNs are leveraged for each dual MDPs. 
The loss function of $\mathcal{L}_{DQN}$ is expressed as
\vspace{-1mm}
\begin{equation}
\mathcal{L}_{DQN} = (Q(s_t,a_t|\theta) - (r_t + \gamma\max_{a_{t+1}}Q(s_{t+1}, a_{t+1}| \theta^\text{Target})))^2,
\label{equ:DQNloss}
\vspace{-2mm}
\end{equation}
where $s_t$, $a_t$, and $r_t$ are the state, action, and reward at timestamp $t$, respectively.
$s_{t+1}$ and $a_{t+1}$ are the state and action at the next timestamp, respectively.
$\gamma$ is the discount factor of the reward.
From the loss function, we update $\theta$, which represents the parameters of DQN. $\theta^\text{Target}$, the target DQN's parameters, are used only for inferencing the Q value of the next state. 
It was first initialized equally with the original DQN. 
Its parameters are updated to the original DQN's parameters after updating the original DQN's parameters ($\theta$) several times.
Fixing the DQN's parameters when estimating the Q value of the next state helps to address the optimization problem caused by the high correlation between sequential data.

The GNN-R model is our \textit{environment} in our dual MDPs because the reward that the environment gives to the agent is related to the score function between the final embedding vectors. 
Thus, optimizing both the GNN-R model and the dual DQN model simultaneously is essential.
In our policy learning, after accumulating positive and negative samples when gaining the reward, the GNN-R model can be optimized by Equation~(\ref{equ:lossfunction}) with those positive and negative pairs.

\begin{algorithm}[t]
\caption{Pseudo-code of Dual Policy learning}\label{alg:cap}

\begin{algorithmic}[1]
\STATE{\textbf{Input}: learning rates of dual DQN ($\alpha_u$ and $\alpha_v$), interacted user-item set $I$, total training epoch $Z$, input graph $\mathcal{G}$, iteration parameter for warm-up $\beta$,  max trajectory length $\kappa$, memory buffer sampling size $b_D$, and \textit{upd} for target DQN update parameter}
\STATE{Initialize parameters: $\theta_u$ of User DQN $Q_u$, $\theta_{u}^\text{Target}$ of target User DQN, $\theta_v$ of Item DQN $Q_v$, $\theta_{v}^\text{Target}$ of target Item DQN, and $\theta_{gnn}$ of GNN-R }

\FOR{$z$= 0, 1, ..., $Z$}
    \STATE{Randomly choose initial user $s_0^u$}
    \STATE{Select initial item $s_0^v$ from $I(s_0^u)$}
    \STATE{$\epsilon = 1- z/Z$}
    \STATE{Initialize two empty lists $L_u$ and $L_v$}
    \FOR{t = 0, 1, ..., $\kappa$}
        \STATE{with probability $\epsilon$, randomly choose $a_t^u$ and $a_t^v$}
        \STATE{Otherwise, acquire $a_t^u$ = $\operatorname*{argmax}_a Q_u(s_t^u, a)$ and $a_t^v$ from $\operatorname*{argmax}_a Q_v(s_t^v, a)$}
        \STATE{Append ($s_{t}^u$, $a_{t}^u$) to $L_u$}
        \STATE{$s_{t+1}^u$ = \textit{NextState}($s_t^u, a_t^u, \mathcal{G}$) via Alg. \ref{alg:next}}
        
        \IF{$t\geq \beta$} 
            \STATE{$r_t^u =$ \textit{Reward}($s_t^u, a_t^u, L_v, I, \theta_{gnn}$) via Alg. \ref{alg:reward}}
            \STATE{Save $s_t^u, a_t^u, r_t^u, s_{t+1}^u$ to user memory}
        \ENDIF
        
        \STATE{Append ($s_{t}^v$, $a_{t}^v$) to $L_v$}
        \STATE{$s_{t+1}^v$ = \textit{NextState}($s_t^v, a_t^v$, $\mathcal{G}$) via Alg. \ref{alg:next}}
        
        \IF{$t\geq \beta$} 
            \STATE{$r_t^v$ = \textit{Reward}($s_t^v, a_t^v, L_u, I, \theta_{gnn}$) via Alg. \ref{alg:reward}}
            \STATE{Save $s_t^v, a_t^v, r_t^v, s_{t+1}^v$ to item memory}
        \ENDIF        
        
    \ENDFOR
    \STATE{Sample $b_D$ $(s_k, a_k, r_k)$ tuples from user and item memory}
    \STATE{$\theta_u\leftarrow\theta_u-\alpha_u\nabla \sum_{k=0}^{b_D} (Q(s_k^u, a_k^u|\theta_u) - (r_k^u+\gamma\max_{a^{u}_{k+1}}Q(s^{u}_{k+1},a^{u}_{k+1}|\theta_{u}^{\text{Target}})))^2$}
    \STATE{$\theta_v\leftarrow\theta_v-\alpha_v\nabla \sum_{k=0}^{b_D} (Q(s_k^v, a_k^v|\theta_v) - (r_k^v+\gamma\max_{a^{v}_{k+1}}Q(s^{v}_{k+1},a^{v}_{k+1}|\theta_{v}^{\text{Target}})))^2$}
    \IF{$z \% \text{\textit{upd}} == 0$}
        \STATE{update $\theta_u^\text{Target}\leftarrow\theta_u$,  $\theta_v^\text{Target}\leftarrow\theta_v$}
    \ENDIF
    \STATE{Train GNN-R model with $L_u$ and $L_v$ to update $\theta_{gnn}$}
\ENDFOR    
\end{algorithmic}
\end{algorithm}

Algorithm~\ref{alg:cap} presents the overall training procedure of \mname{}.
\mname{} trains our dual MDP policy for the whole training epoch $Z$.
At each training step, we first initialize the user state $s_0^u$ and item state $s_0^v$ (L~4 and~5 in Alg.~\ref{alg:cap}). 
Randomness parameter $\epsilon$, defined on L~6 in Alg.~\ref{alg:cap}, determines whether the actions are chosen randomly or from our DQNs (L~9 and~10 in Alg.~\ref{alg:cap}).
\mname{} traverses neighboring nodes and gets rewards until $t$ reaches the max trajectory length $\kappa$ (L~8-23 in Alg.~\ref{alg:cap}).
Furthermore, at each $t$, the (state, action) tuple pairs are saved in the tuple list $L_u$ or $L_v$ (L~11 or 17 in Alg.~\ref{alg:cap}).
The function \textit{NextState} takes the current (state, action) tuple as the input and determines the following user or item states (L~12 or 18 in Alg.~\ref{alg:cap}).
The chosen actions are evaluated using the \textit{Reward} function when $t$ exceeds the warm-up period $\beta$. (L~14 or 20 in Alg.~\ref{alg:cap}).
We note that the quadruple (state, action, reward, and next state) is constructed and saved in the user and item memories.

Finally, User and Item DQN parameters ($\theta_u, \theta_v$) are optimized with the stochastic gradient method (L~25,~26 in Alg.~\ref{alg:cap}).
We update the parameters of target DQNs every target DQN update parameter, \textit{upd} (L~27 and ~28 in Alg.~\ref{alg:cap}). 
The GNN-R model is also trained with positive and negative pairs between users and items, which can be sampled from $L_u$ and $L_v$ (L~30 in Alg.~\ref{alg:cap}). 
Furthermore, if possible, we update the GNN-R parameters by applying BPR loss on target items with positive and negative users, whereas most existing models focus on the loss between target users with positive and negative items.

\subsection{Time Complexity Analysis}
The \textit{NextState} function in Algorithm~\ref{alg:next} leverages a subgraph using the given state and action values. The time complexity of subgraph extraction corresponds to $O(|\mathcal{E}|)$, where $\mathcal{E}$ is the set of all edges of the input graph. We note that this part could be pre-processed and indexed in a constant time because of the limited number of nodes and action spaces. 
The time complexity of the \textit{Reward} function in Algorithm~\ref{alg:reward} is $O(|L|)$ because of the positive and negative tuple generations using the tuple list ($L_u$ or $L_v$). Therefore, the time complexity of a state transition in our MDP (L~8-23 on Alg.~\ref{alg:cap}) is $O(\kappa(|L|+|\mathcal{E}|)) \approx O(|\mathcal{E}|)$ without subgraph extraction pre-processing or $O(\kappa|L|)$ with pre-processing, where $\kappa$ is the maximum trajectory length and a constant.
The time complexity of training the GNN-R model typically corresponds to $O(|\mathcal{E}|)$. In summary, the time complexity of \mname{} is $O(Z|\mathcal{E}|) \approx O(|\mathcal{E}|)$, where $Z$ denotes the total number of training epochs. 

\subsection{Plugging into GNN Recommendation Model}

Our model~\mname{} can be plugged into both existing non-KG-based and KG-based models by slightly modifying the recommendation process.
We aggregate messages the same as existing GNN-R models for the action number of layers.
Furthermore, the pooling process is identical if the dimension of the embedding does not change with respect to different actions, such as in summation-based pooling.
However, in the case of concatenation-based pooling, the dimensions of the final embeddings obtained from different actions are different. To ensure that the scores can be calculated using Equation~(\ref{equ:score}), the dimensions must be equal. To address this issue, we perform an additional padding process to prevent aggregating neighbors beyond the determined number of hops.


\section{Related Work}

\paragraph{Aggregation Optimization for GNNs.}
Aggregation optimization determines the appropriate number of aggregation layers for each node on its input graph, and related studies~\cite{policy-gnn,policygnn2} have recently been proposed. 
For example, Policy-GNN~\cite{policy-gnn} leverages Reinforcement Learning~(RL) to find the policy for choosing the optimum number of GNN layers.
However, all of these studies focused on node classification tasks~\cite{nc1, nc2}, which are inherently different from recommendation tasks. 
Therefore, it is not directly applicable to recommendation tasks, and we also verify whether it has to be a Dual-DQN or Policy-GNN like a single DQN in our ablation study. (refer to Section \ref{subsec:effect of Dual DQN}.)

\begin{table}[t]
\centering
\caption{Comparison to recent GNN-R models} 
\vspace{-2 mm}
\label{t2}
            \resizebox{\linewidth}{!}{
            \begin{tabular}{l|cccc}
            \hline\hline
        & LightGCN~\cite{LightGCN}  & CGKR~\cite{CGKR} & KGIN~\cite{kgin} & \textbf{\mname{}}\\
            \hline
            {High-order Aggregation (Agg.)} & O & O & O & O\\
            {User-aware Agg.} & X & X & O & O\\
            {User/Item-aware Agg. }  & X & X & X & O\\
            {Multi-GNNs and KG/Non-KG Supports} & X & X & X & O\\  
            \hline
            \hline
            \end{tabular}
            }
        \hrule height 0pt
    \vspace{-4mm}
\end{table}

\paragraph{GNNs-based Recommendation (GNN-R)}
GNNs-based recommendation  (e.g., \cite{NGCF, LightGCN, DGCF}) leverages message passing frameworks to model user-item interactions.
GNN-R models stack multiple layers to capture the relationships between the users and items.
GNN-based models have been proposed for learning high-order connectivity~\cite{NGCF, LightGCN}.
Moreover, recent GNN-R models~\cite{DGCF} have adopted different but static aggregation and pooling strategies to aggregate messages from their neighbors.
Another line of research is about utilizing a Knowledge Graph (KG)~\cite{CFKG}. 
A KG can provide rich side information to understand spurious interactions and mitigate cold-start problems and many existing approaches~\cite{kgat,kgin} attempt to exploit the KG by introducing attention-based message passing functions to distinguish the importance of different KG relations.
Although they obtained promising results in different recommendation scenarios, they were still limited in leveraging user/item-aware aggregation strategies.

\paragraph{Reinforcement Learning for Recommendation Tasks.}

Reinforcement Learning (RL) has gained attention for both non-KG datasets \cite{DQN_RS1, DQN_RS2} and KG-based datasets \cite{RLpath1, RLpath2} to enhance recommendation tasks.
Non-KG-based recommendation generates high-quality representations of users and items by exploiting the structure and sequential information~\cite{DQN_RS1} or by defining a policy to handle a large number of items efficiently~\cite{DQN_RS2}.
RL is applied to extract paths from KG or to better represent users and items by generation for the KG-based recommendation.
Paths are extracted by considering the user's interest~\cite{RLG1} or by defining possible meta-paths~\cite{meta-path1}.
However, generation-based approaches generate high-quality representations of users and items, which are applied in both non-KG-based and KG-based recommendations.
For example, RL generates counterfactual edges to alleviate spurious correlations~\cite{CGKR} or generates negative samples by finding the overlapping KG's entity~\cite{RLG2}.

\vspace{-2mm}
\paragraph{Comparison to Recent GNN-R models}
Table~\ref{t2} compares our method to recent baselines with respect to high-order aggregation, user-aware aggregation, user/item-aware aggregation, and Multi-GNNs and KG/non-KG supports. 
First, all the models can aggregate messages over high-order neighborhoods by stacking multiple layers.
However, LightGCN and CGKR do not consider user-aware aggregation strategies for understanding complex user-side intentions. Although KGIN was proposed to learn path-level user intents, it does not optimize the aggregation strategies of both users and items. \mname{} can adaptively adjust the number of GNN-R layers to find better representations of users and items through the dual DQN.
Moreover, \mname{} can be attached and evaluated with many existing GNN-R models, including both KG-based and non-KG-based models, allowing the use of multiple GNN architectures.


\section{Experiment}
We conducted extensive evaluations to verify how our \mname{} works and compare state-of-the-art methods on public datasets by answering the following research questions: \textbf{(RQ1)} What are the overall performances to determine the effectiveness of our method?  \textbf{(RQ2)} How do different components (i.e., Dual-DQN and  Sampling method) affect the performance? \textbf{(RQ3)} What are the effects of the distribution of the chosen GNN layers, groups with different sparsity levels, and the changes in reward over epochs

\subsection{Experimental Settings}

\subsubsection{Datasets.} 

We evaluated \mname{} on three KG datasets and three non-KG datasets.
Table~\ref{tab:datastat} in Appendix shows the statistics for all datasets. 
Details about the data preprocessing of each dataset are described in Appendix~\ref{app:dataset}.

\subsubsection{Evaluation Metrics.}
We evaluated all the models with normalized Discounted Cumulative Gain (nDCG) and Recall, which are extensively used in top-K recommendation~\cite{kgat, NGCF}. We computed the metrics by ranking all items that a user did not interact with in the training dataset. In our experiments, we used K=20. 

\subsubsection{Baselines.}
We evaluated our proposed model on non-KG-based(NFM~\cite{NFM}, FM~\cite{FM}, BPR-MF~\cite{bprloss}, LINE~\cite{line}, NGCF~\cite{NGCF}, DGCF~\cite{DGCF}, and
LightGCN~\cite{LightGCN})  and KG-based model(NFM, FM, BPR-MF,  CKFG~\cite{CFKG}, KGAT~\cite{kgat}, KGIN~\cite{kgin}, CGKR~\cite{CGKR}, and KGPolicy~\cite{RLG2}) models.
We note that CGKR and KGPolicy are recent RL-based GNN-R models that still outperform many existing baselines on public datasets. Refer to Appendix~\ref{app:baseline} for the experimental settings for all baseline models.

\subsubsection{Implementation Details}
\label{app:ID}
For the GNN-based recommendation models, we implemented KGAT \cite{kgat} and LightGCN \cite{LightGCN}.
We set the maximum layer number~($|\mathcal{A}|$) to four after a grid search over [2,6].
KG-based models were fine-tuned based on BPR-MF only if the baseline studies used the pre-trained embeddings.
Other details of the implementation are described in Appendix~\ref{app:ID}. 

\subsection{Performance Comparison (RQ1)}

\begin{table}[t]
\caption{Overall performance comparison on non-KG datasets. We note that the base model GNN-R of \mname{} is LightGCN. \% Improvement (Imp.) shows the relative improvement to the base model. 
Bold font indicates the best result, and underlined results are the second-best models.}
\label{tab:nonKG}
\vspace{-4mm}
        \centering
         \resizebox{\linewidth}{!}{
            \begin{tabular}{c|cc|cc|cc}
            \noalign{\smallskip}\noalign{\smallskip}\hline\hline
            \multirow{0}{*}{} & \multicolumn{2}{c|}{Gowalla} & \multicolumn{2}{c|}{MovieLens1M} &
            \multicolumn{2}{c}{Amazon-Book}\\
            \cline{2-7}
                  & nDCG  & Recall & nDCG & Recall & nDCG & Recall \\
            \hline
             NFM~\cite{NFM} & 0.117 & 0.134 & 0.249 & 0.236 & 0.059 & 0.112 \\
             FM~\cite{FM} & 0.126 & 0.143 & 0.269 & 0.253 & 0.084 & 0.167 \\
             BPR-MF~\cite{bprloss} & 0.128 & 0.150 & 0.282 & \underline{0.273} & 0.084 & 0.160 \\ 
             \hline
             LINE~\cite{line} & 0.068 & 0.120 & 0.249 & 0.240 & 0.070 & 0.142  \\
             NGCF~\cite{NGCF} & \underline{0.225} & \underline{0.154} & 0.267 & 0.266 & \underline{0.097} & \textbf{0.189} \\
             DGCF~\cite{DGCF} & 0.130 & 0.152  & 0.276 & 0.270 & 0.083 & 0.172 \\
             LightGCN~\cite{LightGCN} & \underline{0.225} & 0.152 & \underline{0.283} & \textbf{0.278} & 0.093 & \underline{0.187} \\\hline
             Ours & \textbf{0.229} & \textbf{0.157} & \textbf{0.349} &  \textbf{0.278} & \textbf{0.114} & \textbf{0.189} \\
             
            \hline
            \hline
            \% Imp. & 1.78\% & 3.29\% & 23.3\% & 0\% &  22.58\% & 1.07\% \\
             \hline
             \hline
            \end{tabular}
        }
         \vspace{-3 mm}
\end{table}

\subsubsection{Comparison against non-KG-based methods.} We conducted extensive experiments on three non-KG-based recommendation datasets as listed in Table~\ref{tab:nonKG}.
\mname{} performed the best on all three real-world datasets compared to state-of-the-art models. In particular, \mname{} outperforms the second-best baseline (underlined in Table~\ref{tab:nonKG}) on nDCG@20 by 1.78\%, 23.3\%, and 17.5\%  on Gowalla, MovieLens 1M, and Amazon Book, respectively.
Our model acheived the largest performance gain for the MovieLens 1M dataset, which is denser than the other two datasets.
Since more nodes are considered when choosing the next state, more positive samples and hard negative samples can be stored in our tuple lists, which leads to better representations of users and items than the other baselines.
For sparser datasets (Gowalla and Amazon-Book), \mname{} still performs better than graph-based baselines (NGCF, DGCF, LightGCN) by minimizing the spurious noise via the Dual DQN.

\begin{table}[t]
\centering
\caption{Overall performance comparison of knowledge graph-based recommendations. The base model GNN-R of \mname{} is KGAT. \% Improvement (Imp.) shows the relative improvement to the base model. Bold font indicates the best result, and the second-best model results are underlined.}
\label{tab:KG}
\vspace{-4mm}
        \centering
        \resizebox{\linewidth}{!}{
            \begin{tabular}{c|cc|cc|cc}
            \noalign{\smallskip}\noalign{\smallskip}\hline\hline
            \multirow{0}{*}{} & \multicolumn{2}{c|}{Last-FM} & \multicolumn{2}{c|}{Amazon-Book} &
            \multicolumn{2}{c}{Yelp2018}\\
            \cline{2-7}
                  & nDCG  & Recall & nDCG & Recall & nDCG & Recall \\
            \hline
             FM~\cite{FM} & 0.0627 & 0.0736 & 0.2169 & 0.4078 & 0.0206 & 0.0307 \\
             NFM~\cite{NFM} & 0.0538 & 0.0673 & 0.3294 & \underline{0.5370} & 0.0258 & 0.0418 \\
             BPR-MF~\cite{bprloss} & 0.0618 & 0.0719 & 0.3126 & 0.4925 & 0.0324 & 0.0499\\
             \hline

             CKFG~\cite{CFKG} & 0.0477 & 0.0558 & 0.1528 & 0.3015 & 0.0389 & 0.0595 \\
             KGAT~\cite{kgat} & 0.0823 & 0.0948 & 0.2237 & 0.4011 & 0.0413 & 0.0653\\
             KGIN~\cite{kgin} & \underline{0.0842} & 0.0972 & 0.2205 & 0.3903 & \underline{0.0451} & \textbf{0.0705} \\ \hline

            CGKR~\cite{CGKR} & 0.0660 & 0.0750 & \underline{0.3337} & 0.5186 & 0.0372 & 0.0579 \\ 
             KGPolicy~\cite{RLG2} & 0.0828 & \underline{0.0978} & 0.3316 & 0.5286 & 0.0384 & 0.0509 \\\hline
             Ours & \textbf{0.0889} & \textbf{0.1072} & \textbf{0.3662} & \textbf{0.5735} & \textbf{0.0452} & \underline{0.0700} \\\hline
            \hline
            \% Imp. & 8.02\% & 13.1\% & 63.7\% & 42.9\% & 9.44\% & 7.20\% \\
             \hline
             \hline
            \end{tabular}
  }
        \vspace{-3mm}
\end{table}

\begin{figure*}[t]
  \centering
  \begin{subfigure}[b]{0.24\textwidth}
    \centering
    \includegraphics[width=\textwidth, height=2.2cm]{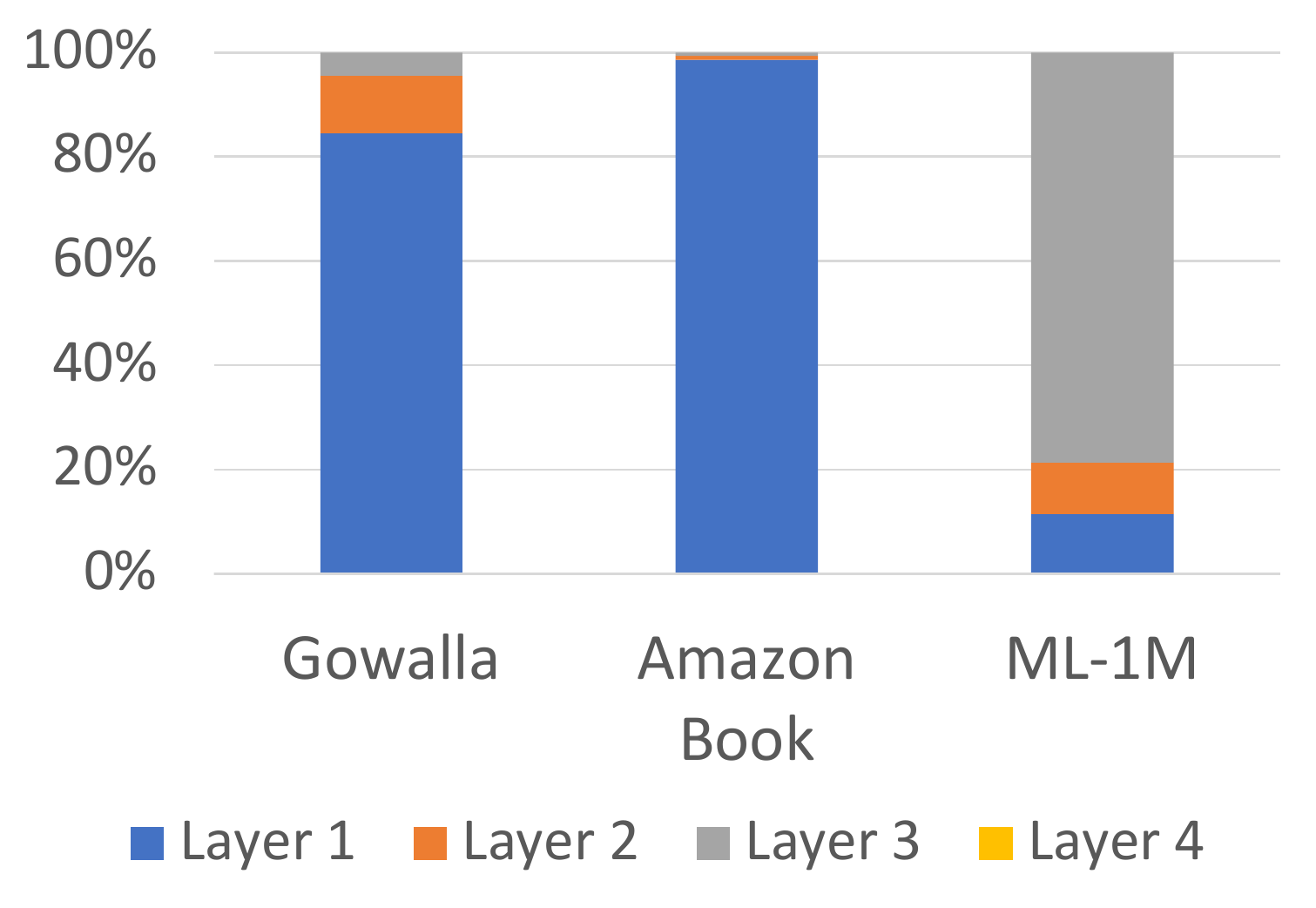}
    \caption{Users in Non-KG}
    \label{fig:nonuser}
  \end{subfigure}
  \begin{subfigure}[b]{0.24\textwidth}
    \centering
    \includegraphics[width=\textwidth, height=2.2cm]{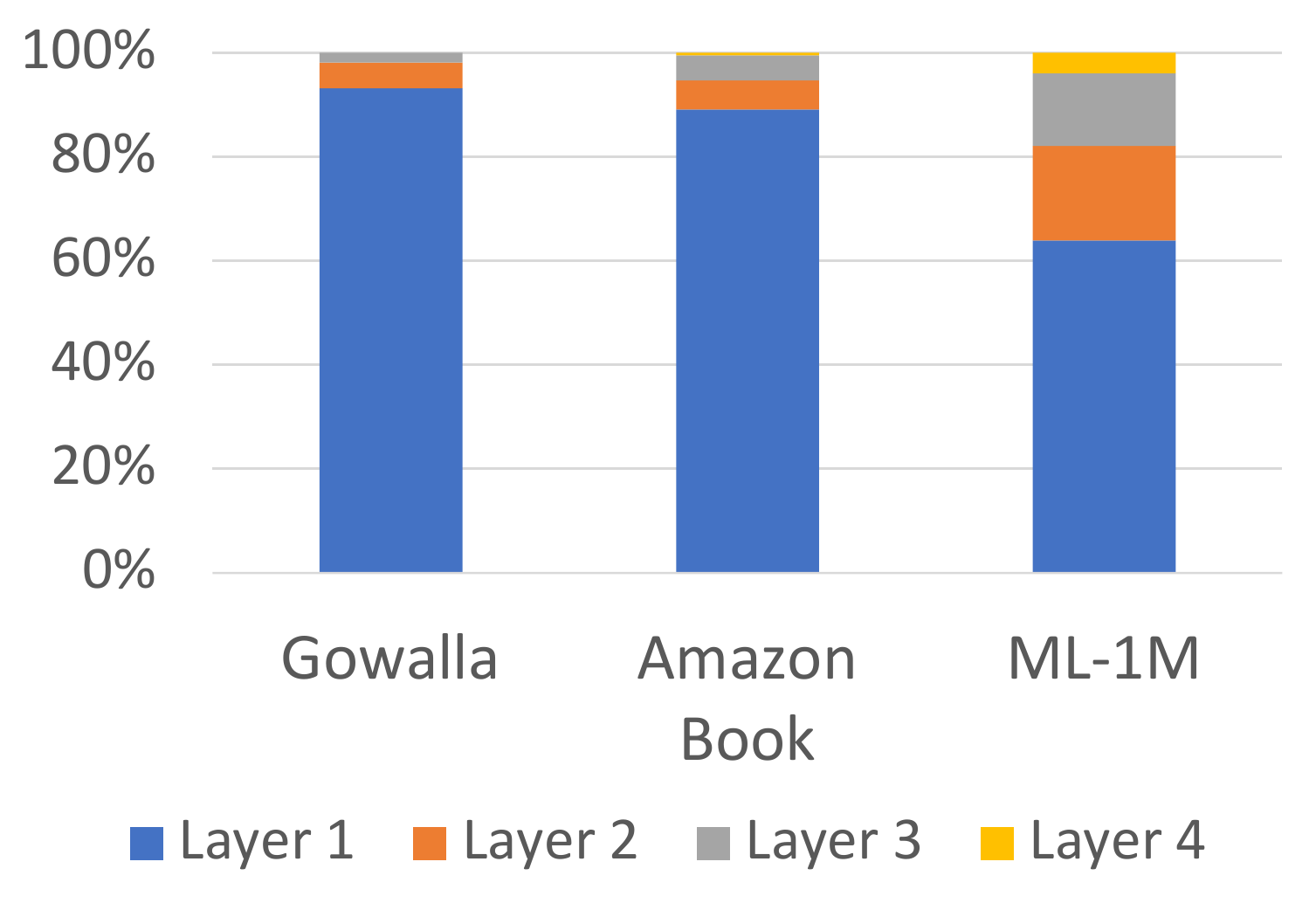}
    \caption{Items in Non-KG}
    \label{fig:nonitem}
  \end{subfigure}
  \begin{subfigure}[b]{0.24\textwidth}
    \centering
    \includegraphics[width=\textwidth, height=2.2cm]{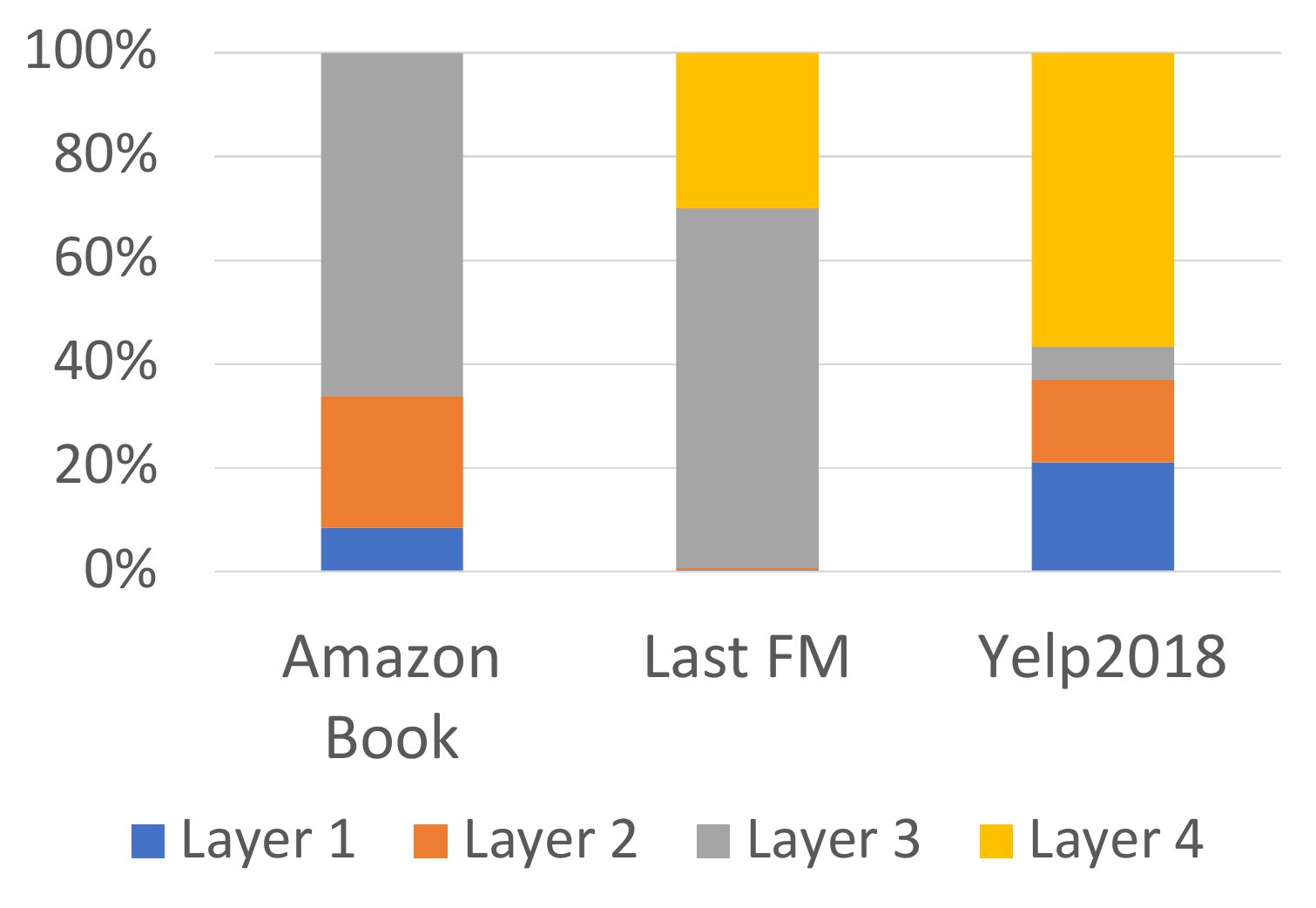}
    \caption{Users in KG}
    \label{fig:KGuser}
  \end{subfigure}
  \begin{subfigure}[b]{0.24\textwidth}
    \centering
    \includegraphics[width=\textwidth, height=2.2cm]{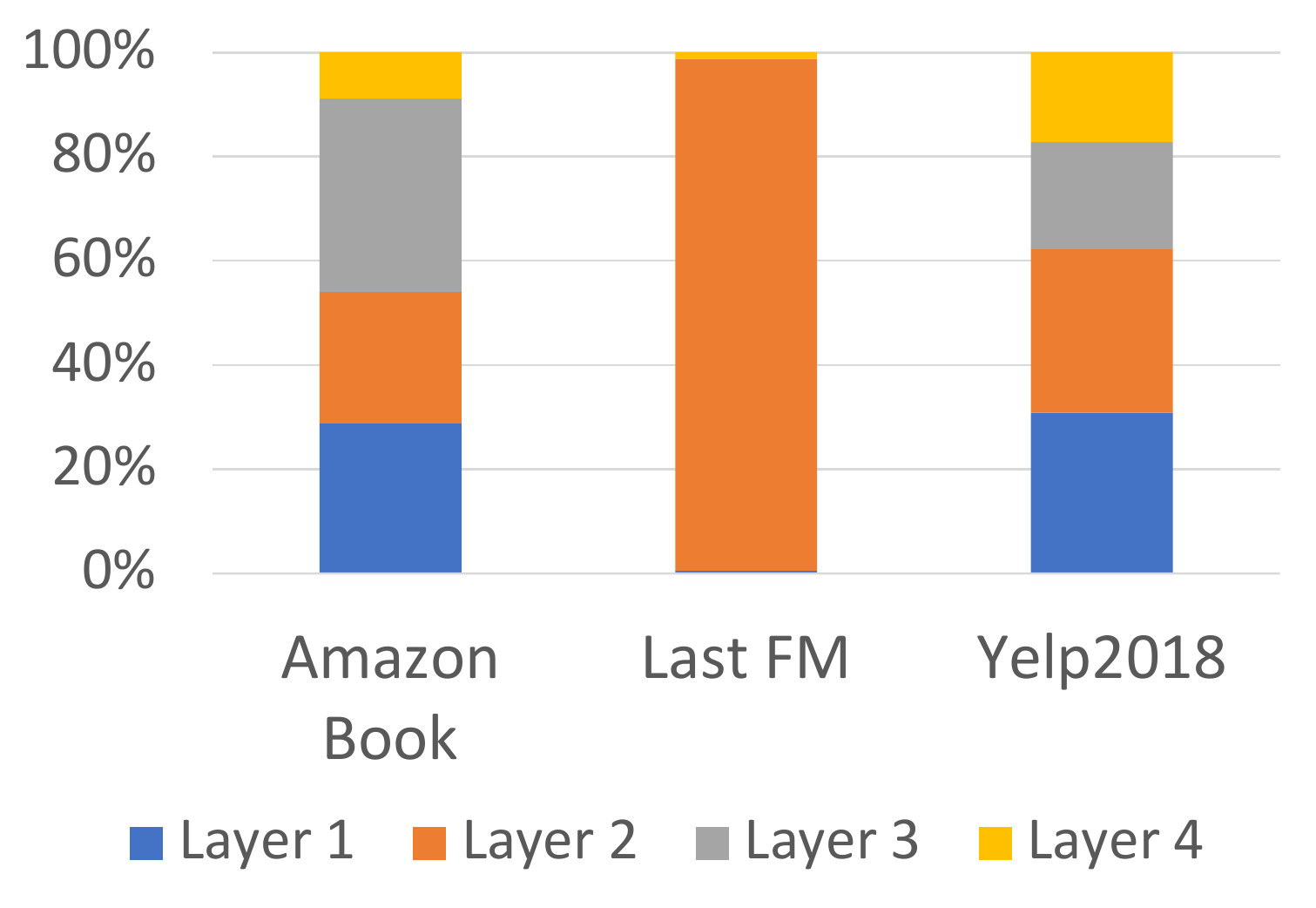}
    \caption{Items in KG}
    \label{fig:KGitem}
  \end{subfigure}
 \vspace{-2 mm} 
  \caption{Distribution of the chosen actions (i.e., the number of layers used to obtain user and item embeddings) in \mname{}. In each subplot, the distribution of the assigned actions is shown with a percentage. Blue, red, gray, and yellow indicate action 1, 2, 3, and 4, respectively.}
  \label{fig:layerdist}
  \vspace{-2 mm} 
\end{figure*}

\subsubsection{Comparison against KG-based methods.} We also experimented with \mname{} on KG-based recommendation models. 
Table~\ref{tab:KG} presents a performance comparison of KG-based recommendations. 
While NFM obtained better results than a few other high-order-based models in  Amazon-Book, graph-based models (CKFG, KGAT, KGIN, CGKR, and KGPolicy) generally achieved better performance than the matrix factorization-based models (FM, NFM, and BPR-MF). 
KGPolicy and CGKR, based on RL, showed good performance compared with the other baselines. 
CGKR generates counterfactual interactions, and KGPolicy searches high-quality negative samples to enhance their GNN-Rs. However, DPAO minimizes spurious noise by adaptively determining the GNN layers and finding the hard positive and negative samples to determine the best GNN layers. 
Therefore, \mname{} achieved the best results for the nDCG and Recall measures for most datasets. Our model achieved 7.37\% and 15.8\% improvements on nDCG, and 9.61\%, 10.6\% improvements on Recall on Last FM and Amazon Book, respectively, compared with the second best model.
This enhancement implies that the proposed RL leverages KGs more effectively.
While KGIN shows worse results for Last-FM and Amazon-Book, it produces comparable performances to our \mname{} on Yelp2018 dataset.
KGIN has the advantage of capturing path-based user intents. This demonstrates that item-wise aggregation optimization may not be effective on the Yelp dataset. 

\subsection{Ablation study (RQ2)}

\begin{table}
\centering
\caption{Ablation study of Dual DQN and sampling in our method~\mname{}.}
\vspace{-4mm}
\label{tab:1DQN}

\resizebox{\linewidth}{!}{
\scalebox{0.9}{
\begin{tabular}{c|cc|cc|cc}
\noalign{\smallskip}\noalign{\smallskip}\hline\hline
 \multirow{0}{*}{}& \multicolumn{2}{c|}{\mname{}} & \multicolumn{2}{c|}{ w/o Dual} & \multicolumn{2}{c}{w/o sampling}\\
\cline{2-7}
      & nDCG& Recall& nDCG &Recall &nDCG &Recall \\
\hline
Amazon Book & 0.3662 & 0.5735 & 0.3491 & 0.5503 &  0.3422 & 0.5148 \\
Last FM & 0.0889 & 0.1072 & 0.0876 & 0.0975 & 0.0876 & 0.0975  \\
Gowalla  & 0.229 & 0.157 & 0.228 & 0.155 & 0.226 & 0.155 \\
MovieLens 1M & 0.349 & 0.278 & 0.335 & 0.263 & 0.341 & 0.257 \\
\hline
\hline
\end{tabular}
}
}

\vspace{-4mm}
\end{table}

\subsubsection{Effect of Dual DQN}
\label{subsec:effect of Dual DQN}
Table~\ref{tab:1DQN} shows the effectiveness of dual DQN by comparing the \mname{} with the Single DQN. The Single DQN exploits the same DQN model for User DQN and Item DQN.
We observed that the Single DQN decreases the recommendation performances on KG datasets (Amazon Book and Last FM) and non-KG datasets (Gowalla and MovieLens1M).
The performance gains were greater for KG datasets using the proposed dual DQN.
This result indicates that it is more effective to develop heterogeneous aggregation strategies for users and items linked to KGs.

\subsubsection{Effect of Sampling}
\label{subsec:effect of sampling}

To verify the effectiveness of our sampling technique when computing \textit{Reward} in Algorithm \ref{alg:reward}, we substitute our method to Random Negative Sampling (RNS)~\cite{bprloss}. RNS randomly chooses users (or items) from all users (or items) for negative sampling.
As shown in Table~\ref{tab:1DQN}, our proposed sampling technique outperformed random sampling.
Our sampling approach obtains samples from closer neighbors, resulting in improved nDCG and Recall scores.
Moreover, the time complexity of our sampling method depends on the size of tuple lists, which is much less than the number of all users (or items.) 

\subsection{Further Analysis (RQ3)}
\subsubsection{Analysis of Layer Distribution}
To ensure that each user and item adaptively choose a different action, we analyzed the layer distributions obtained from the dual DQN.
Fig.~\ref{fig:layerdist} visualizes the distributions of layer selections (i.e., actions) returned from \mname{}.
We analyzed the distribution of the chosen actions for users in non-KG, non-KG items, KG users, and KG items.
As shown in Figure~\ref{fig:layerdist}, the distribution of layers are different for distinct datasets.
These results further confirm that optimizing the aggregation strategy for each user and item is important.

An interesting observation is that a single layer is frequently selected for non-KG-based datasets.
The selection of a single layer is reasonable because a direct collaborative signal from a 1-hop neighbor gives as much information.
Furthermore, we observe that the user and item distribution are different, especially for the KG datasets.
Conversely, two or more layers are selected more frequently from the KG-based datasets. 
Users in KG-based datasets need more than one layer to obtain information from the KG.

\begin{figure}[!t]
  \centering
  \begin{subfigure}[b]{0.32\linewidth}
    \centering
    \includegraphics[width=\linewidth]{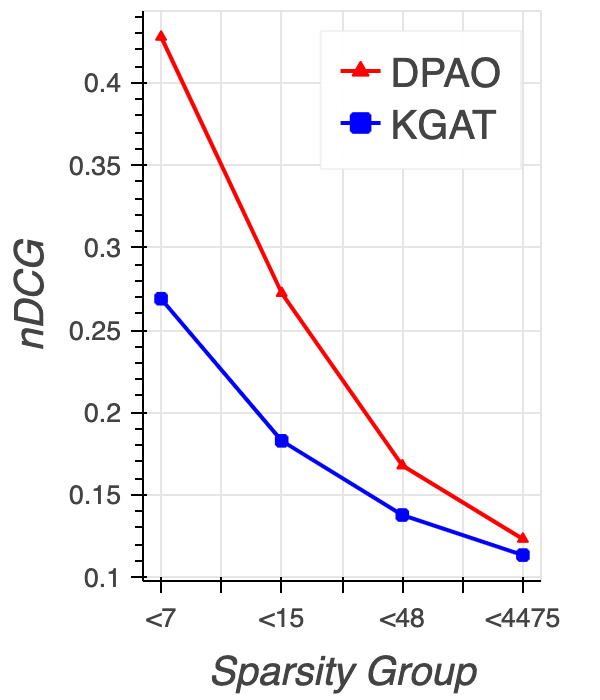}
    \caption{Amazon-Book}
    \label{fig:ab-sparse}
  \end{subfigure}
  \hfill
  \begin{subfigure}[b]{0.32\linewidth}
    \centering
    \includegraphics[width=\linewidth]{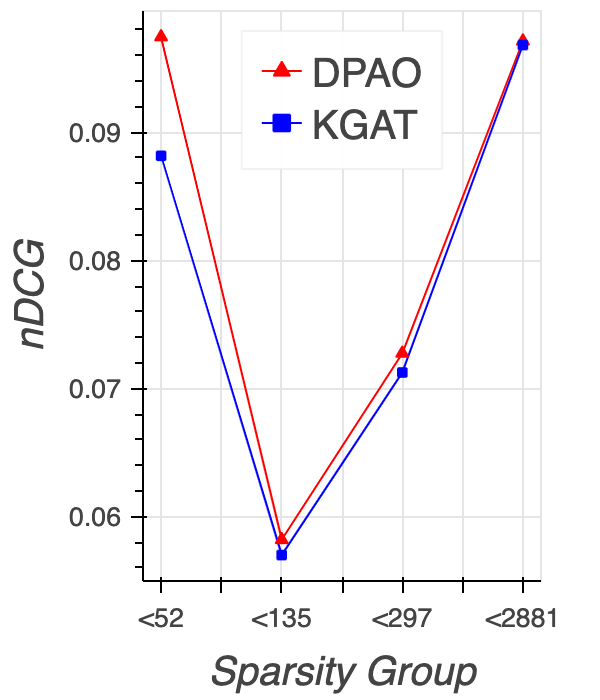}
    \caption{Last-FM}
    \label{fig:lf-sparse}
  \end{subfigure}
  \hfill
  \begin{subfigure}[b]{0.32\linewidth}
    \centering
    \includegraphics[width=\linewidth]{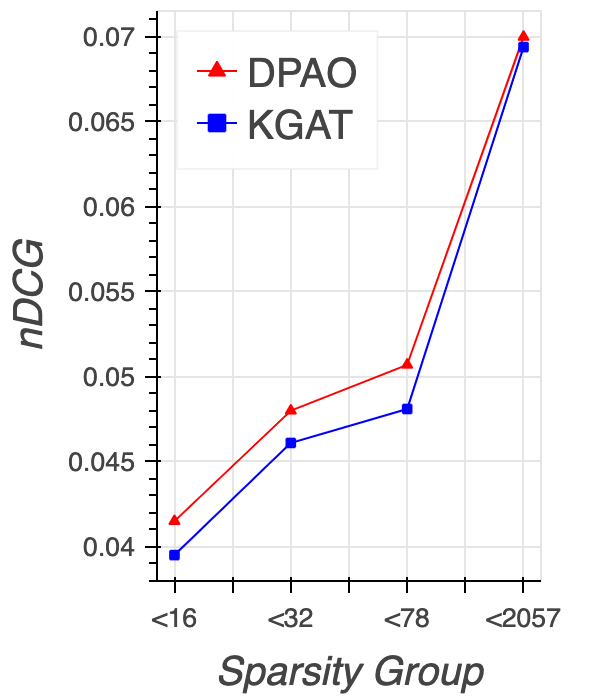}
    \caption{Yelp2018}
    \label{fig:yelp-sparse}
  \end{subfigure}
  \vspace{-2mm} 
  \caption{Performance comparison of different datasets over different sparsity levels. nDCG@20 was used to check the performance.}
  
  \label{fig:sparse}
  \vspace{-4mm}
\end{figure}

\subsubsection{Analysis with respects to Sparsity}
Finally, we examined the performance differences over different sparsity levels.
We divided all users into four groups, and the nDCG was calculated for each group.
For the Amazon-Book dataset, interaction numbers less than 7, 15, 48, and 4475 are chosen to classify the users. These numbers were selected to keep the total interactions for each group as equal as possible by following \cite{kgat}.
Fig~\ref{fig:sparse} shows the performance of \mname{} and our base model, KGAT, with respect to different sparsity levels.

\mname{} outperformed the baseline in every sparsity group. 
Our model improves the performance of KGAT by reducing the spurious noise that may arise from fixing the number of layers.
Furthermore, we can observe that \mname{} shows the largest performance gain for the smallest interaction group on the Amazon-Book and Last-FM datasets.
This result indicates that it is more important for the users with fewer interactions to choose the aggregation number adaptively.

\subsubsection{Change of Rewards over Epochs}
The reward function of \mname{} is defined in Equation~(\ref{equ:reward}).
Higher rewards mean that positive and negative boundaries become apparent, which in turn means better recommendations.
Figure~\ref{fig:reward} in Appendix shows the change in rewards over time for the non-KG datasets.
The numerator part of the reward is plotted on the Figure~\ref{fig:reward} to better visualize the amount of change.
One notable observation in Figure~\ref{fig:reward} is the high fluctuation in reward.
A reasonable reason for the reward fluctuation is the stochastic selection of negative samples.
However, as we can observe from the trend line, the average value of each epoch, the reward, increases as the epoch proceeds.
In summary, the reward increases over time, indicating that \mname{} effectively optimizes the representations of users and objects by adaptively selecting appropriate actions.


\section{Acknowledgement}
This work was supported by the National Research Foundation of Korea (NRF) (2021R1C1C1005407, 2022R1F1A1074334). In addition, this work was supported by the Institute of Information \& communications Technology Planning \& evaluation (IITP) funded by the Korea government (MSIT):  (No. 2019-0-00421, Artificial Intelligence Graduate School Program (Sungkyunkwan University) and (No. 2019-0-00079, Artificial Intelligence Graduate School Program (Korea University)).

\section{Conclusion}
Our framework proposes an aggregation optimization model that adaptively chooses high-order connectivity to aggregate users and items.
To distinguish the behavior of the user and item, we defined two RL models that differentiate their aggregation distribution.
Furthermore, our model uses item-wise and user-wise sampling to obtain rewards and optimize the GNN-R model.
Our proposed framework was evaluated on both non-KG-based and KG-based GNN-R models under six real-world datasets. Experimental results and ablation studies show that our proposed framework significantly outperforms state-of-the-art baselines.

\newpage

\bibliographystyle{ACM-Reference-Format}
\bibliography{ref}

\newpage
\appendix

\section{Experimental setup}

\begin{figure}[!b]
  \centering
    
  \begin{subfigure}[b]{0.32\linewidth}
    \centering
    \includegraphics[width=\linewidth]{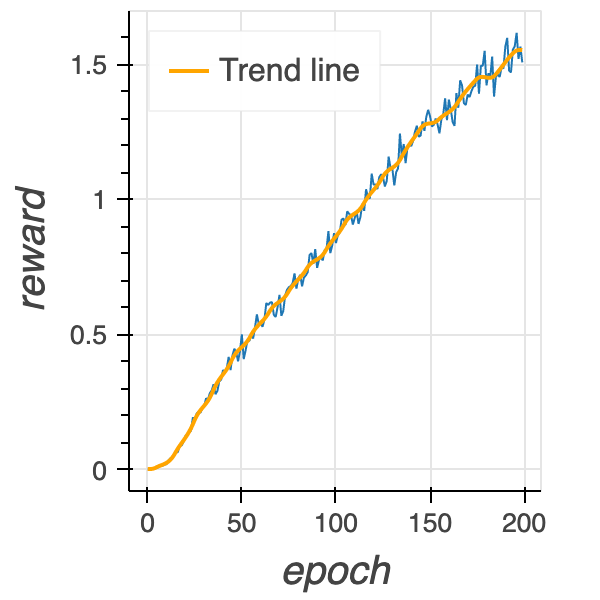}
    \caption{ML1M-item}
    \label{fig:ml-item}
  \end{subfigure}
  \hfill
  \begin{subfigure}[b]{0.32\linewidth}
    \centering
    \includegraphics[width=\linewidth]{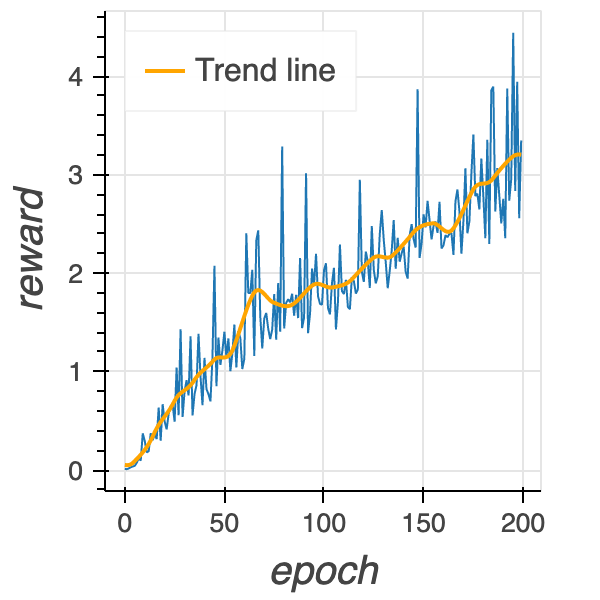}
    \caption{Amazon-item}
    \label{fig:ab-item}
  \end{subfigure}
  \hfill
  \begin{subfigure}[b]{0.32\linewidth}
    \centering
    \includegraphics[width=\linewidth]{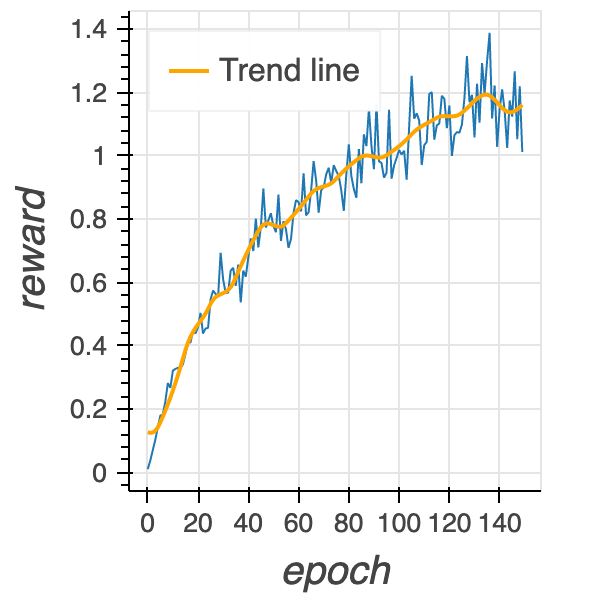}
    \caption{Gowalla-item}
    \label{fig:gowalla-item}
  \end{subfigure}
  \begin{subfigure}[b]{0.32\linewidth}
    \centering
    \includegraphics[width=\linewidth]{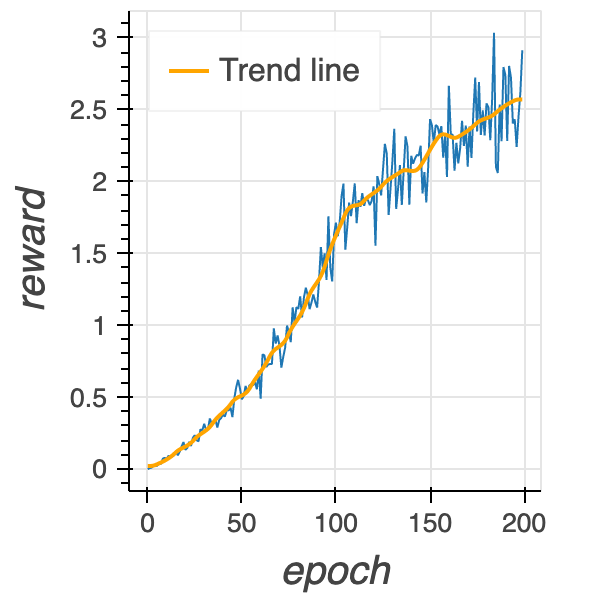}
    \caption{ML1M-user}
    \label{fig:ml-user}
  \end{subfigure}
  \hfill
  \begin{subfigure}[b]{0.32\linewidth}
    \centering
    \includegraphics[width=\linewidth]{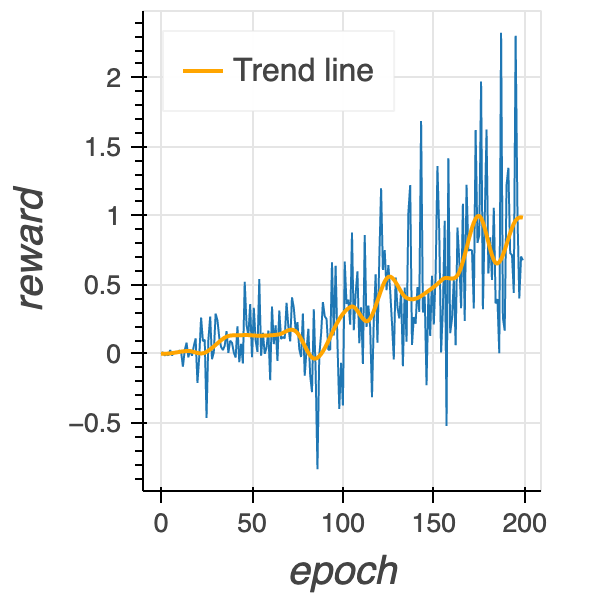}
    \caption{Amazon-user}
    \label{fig:ab-user}
  \end{subfigure}
  \hfill
  \begin{subfigure}[b]{0.32\linewidth}
    \centering
    \includegraphics[width=\linewidth]{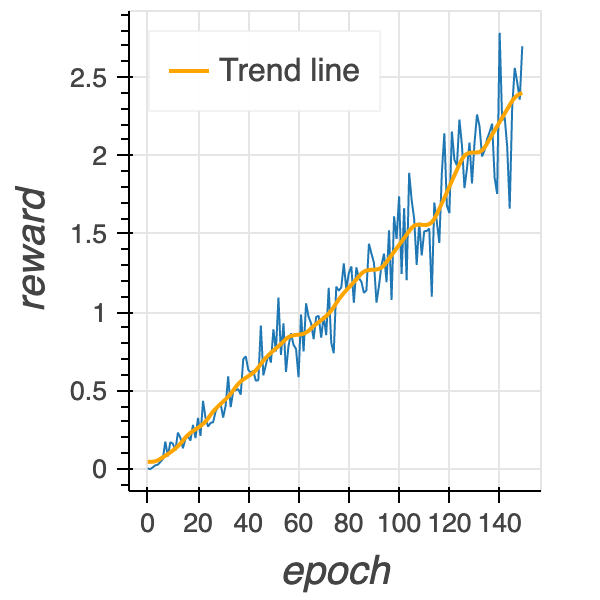}
    \caption{Gowalla-user}
    \label{fig:gowalla-user}
  \end{subfigure}
  \caption{Reward curve of our \mname{}. The first row indicates the item DQN's reward and the second row expresses the user DQN's reward. The trend line colored in orange represents the overall average of the reward. }
  \label{fig:reward}
\end{figure}

\begin{table}[b]
\centering
\caption{Statistics of non-KG-based and KG-based datasets}
\label{tab:datastat}
\scalebox{0.9}{
\resizebox{\linewidth}{!}{
\begin{tabular}{c|ccc}
\noalign{\smallskip}\noalign{\smallskip}\hline\hline
& Gowalla & Amazon Book & MovieLens 1M \\
\hline
\#Users & 29,858 & 70,679 & 6,040 \\
\#Items & 40,981 & 24,915 & 3,953 \\
\#Interactions  & 1,027,370 & 846,703 & 1,000,017 \\

\hline
\noalign{\smallskip}\noalign{\smallskip}\hline

& Last FM  & Amazon Book & Yelp2018 \\
\hline
\#Users & 23,566 & 70,679 & 45,919 \\
\#Items & 48,123 & 24,915 & 45,538 \\
\#Interactions  & 3,034,796 & 846,703 & 1,183,610 \\
\#Relations & 9 & 39 & 42 \\
\#Entities & 58,134 & 88,572 & 90,499 \\
\#Triplets & 464,567 & 2,557,746 & 1,853,704 \\
\hline

\hline
\end{tabular}
}
}
\end{table}

\subsection{Datasets}
\label{app:dataset}

\hspace{\parindent} \textbf{Gowalla\footnotemark{},\footnotetext{https://snap.stanford.edu/data/loc-gowalla.html}} is a social networking network dataset that uses location information, where users share their locations by checking in. We are strict in that user and item interactions must have more than 10 interactions.

\textbf{MovieLens1M\footnotemark{}, \footnotetext{https://grouplens.org/datasets/movielens/}} is a movie dataset that consists of ratings of movies. Each person expresses a preference for a movie by rating it with a score of 0–5. The preference between the user and the movie is defined as implicit feedback and is connected as edges.

\textbf{Amazon-book\footnotemark{}. \footnotetext{http://jmcauley.ucsd.edu/data/amazon} } is a book dataset in Amazon-review used for product recommendation.

\textbf{Last-FM\footnotemark{} \footnotetext{http://millionsongdataset.com/lastfm/\#getting}} is a music dataset derived from Last.fm online music systems. In this experiment, we used a subset of the dataset from Jan 2015 to June 2015.

\textbf{Yelp2018\footnotemark{} \footnotetext{https://www.yelp.com/dataset/download}} is a dataset from the Yelp challenge in 2018. Restaurants and bars were used as items.  

In addition to the user-item interaction, we need to construct a KG for each dataset~(Last-FM, Amazon-Book, Yelp2018).
For Last-FM and Amazon-Book, we mapped the items into entities in Freebase. 
We used an information network, such as location and attribute, as an entity of KG in the Yelp2018 dataset.

For each dataset, we randomly selected 80\% of the interaction set of each user to create the training set and regarded the remainder as the test set. From the training set, we randomly selected 10\% of the interactions as a validation set to tune the hyperparameters. Moreover, we used the 10-core setting for every user-item interaction dataset.
For the Amazon Book dataset, to eliminate the effect of co-existing interactions in the training and test dataset, we split the dataset with the same split ratio with no duplicate interactions on the training and test dataset.

\subsection{Baseline}
\label{app:baseline}

We evaluated our proposed model on non-KG-based and KG-based model.

\textbf{Non-KG-based: }
To demonstrate the effectiveness of our method, we compared our model with supervised learning methods(NFM, FM, BPR-MF) and graph-based models(LINE, LightGCN, NGCF, DGCF):
\begin{itemize}
    \item \textbf{FM} \cite{FM} is a factorization model that considers the feature interaction between inputs. It only considers the embedding vector itself to perform the prediction.
    \item \textbf{NFM} \cite{NFM} is also a factorization model. Its difference from the FM model is that it uses neural networks to perform prediction. We employed one hidden layer.
    \item \textbf{BPR-MF} \cite{bprloss} is a factorization model that optimizes using negative pairs. They utilize pairwise ranking loss. 
    \item \textbf{LINE} \cite{line} is a model that uses sampling to apply to both big and small data.
    \item \textbf{NGCF} \cite{NGCF} is a propagation based GNN-R model. The model transforms the original GCN layer by applying the element-wise product in aggregation.
    \item \textbf{LightGCN} \cite{LightGCN} is a state-of-the-art GNN-R model. The model simplifies the NGCF structure by reducing learnable parameters.
    \item \textbf{DGCF} \cite{DGCF} is a GNN-R model that gives a high signal on the collaborative signal by giving latent intents.
\end{itemize}

\textbf{KG-based: }
For the baseline in the KG-based recommendation, we compared supervised learning (NFM, FM, BPR-MF), regularized (CFKG), graph-based (KGAT, KGIN) model, and RL-based (KGPolicy).
\begin{itemize}
    \item \textbf{FM, NFM} is the same model discussed in the non-KG-based models. The difference is that we used the KG as input features. We used the knowledge connected directly to its user (item).
    \item \textbf{CFKG} \cite{CFKG} applies TransE on the CKG, which includes users, items, and entities. This model finds the interaction pair with the plausibility prediction.
    \item \textbf{KGAT} \cite{kgat} uses the GAT layer at aggregation. The model aggregates on CKG. It applies an attention mechanism to relations to distinguish the importance of relations based on users.
    \item \textbf{KGIN} \cite{kgin} is a model that gives a higher collaborative signal on interaction set than other KG relations. It formulates extra relations called intents on user items to give a high signal on interaction.
    \item \textbf{KGPolicy} \cite{RLG2} is a model that finds negative samples with RL. They choose a negative sample by receiving knowledge-aware signals. This model is performed on the FM model by adapting KGPolicy negative sampling techniques.
    \item \textbf{CGKR}~\cite{CGKR} is a model that generates spurious correlations on entities on KG that utilizes Reinforcement Learning.
\end{itemize}
\begin{table}
\centering
\caption{Total training time comparison between \mname{} and base model on the non-KG dataset (in minutes)}
\label{tab:time-nonKG}
\resizebox{\linewidth}{!}{
\scalebox{1.0}{
\begin{tabular}{c|ccc}
\noalign{\smallskip}\noalign{\smallskip}\hline\hline
 \multirow{0}{*}{}& \multicolumn{3}{c}{{non-KG Datasets}}  \\
\cline{2-4}
    & Gowalla &MovieLens1M &Amazon-Book \\
\hline
\mname{}&  40.23 &  41.58 & 33.33 \\
LightGCN~\cite{LightGCN}  & 61.66 & 91.6 & 51.67  \\
\hline
\hline
\end{tabular}
}
}
\end{table}

\begin{table}
\centering
\caption{Total training time comparison between \mname{} and base model on the KG dataset (in minutes)}
\label{tab:time-KG}

\resizebox{\linewidth}{!}{
\scalebox{1.0}{
\begin{tabular}{c|ccc}
\noalign{\smallskip}\noalign{\smallskip}\hline\hline
 \multirow{0}{*}{}& \multicolumn{3}{c}{{KG Datasets}} \\
\cline{2-4}
    & Last FM & Amazon-Book & Yelp2018 \\
\hline
\mname{}&  256.66 &  420.71 & 433.33 \\
KGAT~\cite{kgat} & 324 & 468.3   & 350  \\
\hline
\hline
\end{tabular}
}
}

\end{table}
\subsection{Implementation Details}
\label{app:ID}
We constructed multiple Multi-Layer Perceptrons (MLPs) to construct our DQN model. 
The SGD optimizer~\cite{SGD} was used to train the two DQN models with an initial learning rate as 0.001 and discount factor $\gamma$ as 0.98.
Hyperparameters not stated above were tuned with a grid search.
The maximum length of the trajectory was tuned amongst~\{10, 20, 40, 100\}. A number of trajectories were chosen from \{10, 20, 30, 100\}.
A number of timestamps were tuned within \{100, 150, 200\}.
For the KGAT training, we used 64 as the initial embedding dimension.
The Adam optimizer was used to optimize the KGAT model.
The epochs were set to 1000, and we established early stopping when the validation performance did not increase for 10 steps.
The other hyperparameters were tuned using a grid search;
learning rates were found with \{0.01, 0.001, 0.0001, 0.00001\}.
The batch size was chosen as \{512, 1024, 2048\}.
All dimension of the 4 (max) layer were tuned amongst \{16, 32, 64, 128, 256\}.
The dropout of these layers was tuned to \{0, 0.1, 0.2, 0.4, 0.5, 0.6\}.
For LightGCN optimization, we used 64 as the initial embedding dimension along with the Adam optimizer.
Here, 400 is the number of epochs, and the same early stopping was used.
The dropout and batch sizes were set as 0.1 and 1024 in our experiment, respectively.
The learning rates were tuned by a grid search in \{0.01, 0.001, 0.0001, 0.00001\}.
Finally, for other baseline models, we followed the original papers for the hyperparameters settings and used a grid search if the dataset was not used in the original papers.

\section{(Extended) Experimental Results}

\subsection{Total training time comparison}
We compared the training time of \mname{} with those of other baselines.
Table~\ref{tab:time-nonKG} and ~\ref{tab:time-KG} present the amount of time spent on the non-KG datasets and KG datasets, respectively.
For \mname{}, we used the pre-processed subgraph extraction when transitioning to the next state.
As shown in Tables~\ref{tab:time-nonKG} and ~\ref{tab:time-KG}, \mname{} is faster than the other GNN-R model.
Compared to the base model, our model required a similar amount of time on the other datasets.
This is because the time complexity for the next state transition and the reward are the same as constant.
The only difference occurs when training GNN-R on \mname{}, which depends on the size of the positive samples.

\end{document}